\pdfoutput=1
\documentclass[11pt]{article}

\usepackage[final]{acl}
\usepackage{times}
\usepackage{latexsym}
\usepackage{colortbl}
\usepackage{xcolor}
\usepackage{soul}
\sethlcolor{red!20}
\usepackage{siunitx}
\usepackage{svg}
\usepackage{bbm}
\usepackage{booktabs}
\usepackage{multirow}
\usepackage{amsmath}
\usepackage{adjustbox}
\usepackage{pgfplots}
\pgfplotsset{compat=1.18} % or another version like 1.17
\usepackage{enumerate}

\usepackage{paralist}
\usepackage{enumitem}
\usepackage{amsmath,amssymb}

\usepackage[normalem]{ulem}
\useunder{\uline}{\ul}{}
\usepackage[T1]{fontenc}
\usepackage[utf8]{inputenc}

\usepackage{microtype}

\usepackage{inconsolata}

\usepackage{graphicx}
\usepackage{fontawesome5}
\usepackage[most]{tcolorbox}
\tcbuselibrary{listings,breakable}
\newtcblisting{tcbverbatim}{
  breakable,
  colback=white,
  colframe=white,
  boxrule=0pt,             % no border line thickness
  listing only,
  listing options={
    basicstyle=\ttfamily\small,
    breaklines=true,           % enable automatic line breaking
    breakatwhitespace=true,    % break lines at whitespace if possible
  },
  left=6pt,
  right=6pt,
  top=6pt,
  bottom=6pt,
  enhanced,
  fonttitle=\bfseries
}

\newtcolorbox{takeawaybox}{
  enhanced,
  breakable,
  colback=black!5!white,        % very soft blue
  colframe=black!10!white,      % soft but noticeable border
  coltitle=black,              % high contrast title
  fonttitle=\bfseries,
  title={\faLightbulb\hspace{0.5em}Takeaway},
  boxrule=0.4pt,
  arc=3pt,
  left=8pt,
  right=8pt,
  top=6pt,
  bottom=6pt,
  before skip=10pt,
  after skip=10pt
}

\title{When Models Lie, We Learn: Multilingual Span-Level Hallucination Detection with PsiloQA}

\author{
 \textbf{Elisei Rykov\textsuperscript{1}},
 \textbf{Kseniia Petrushina\textsuperscript{1,5}},
 \textbf{Maksim Savkin\textsuperscript{2,5}},
 \textbf{Valerii Olisov\textsuperscript{5}},
 \\
 \textbf{Artem Vazhentsev\textsuperscript{2,1}},
 \textbf{Kseniia Titova\textsuperscript{3,1}},
 \textbf{Alexander Panchenko\textsuperscript{1,2}},
\\
 \textbf{Vasily Konovalov\textsuperscript{2,1,5}, \textbf{and}
 \textbf{Julia Belikova\textsuperscript{4,1}}}
\\
\textsuperscript{1}Skoltech,
\textsuperscript{2}AIRI,
\textsuperscript{3}MWS AI,
\textsuperscript{4}Sber AI Lab,\\
\textsuperscript{5}Moscow Institute of Physics and Technology
\\
\href{mailto:elisei.rykov@skol.tech}{\{Elisei.Rykov}, 
\href{mailto:a.panchenko@skol.tech}{A.Panchenko},
\href{mailto:julia.belikova@skol.tech}{Julia.Belikova\}}@skol.tech
}

\begin{document}
\maketitle
\begin{abstract}
Hallucination detection remains a fundamental challenge for the safe and reliable deployment of large language models (LLMs), especially in applications requiring factual accuracy. Existing hallucination benchmarks often operate at the sequence level and are limited to English, lacking the fine-grained, multilingual supervision needed for a comprehensive evaluation. In this work, we introduce PsiloQA, a large-scale, multilingual dataset annotated with span-level hallucinations across 14 languages. PsiloQA is constructed through an automated three-stage pipeline: generating question–answer pairs from Wikipedia using GPT-4o, eliciting potentially hallucinated answers from diverse LLMs in a no-context setting, and automatically annotating hallucinated spans using GPT-4o by comparing against golden answers and retrieved context. We evaluate a wide range of hallucination detection methods -- including uncertainty quantification, LLM-based tagging, and fine-tuned encoder models -- and show that encoder-based models achieve the strongest performance across languages. Furthermore, PsiloQA demonstrates effective cross-lingual generalization and supports robust knowledge transfer to other benchmarks, all while being significantly more cost-efficient than human-annotated datasets. Our dataset and results advance the development of scalable, fine-grained hallucination detection in multilingual settings.\footnote{\url{https://github.com/s-nlp/psiloqa}}
\end{abstract}

\section{Introduction}

Large Language Models (LLMs) became a crucial component in a wide range of text generation applications, including summarization, translation, and question-answering systems in various domains. However, even state-of-the-art models, such as GPT-4~\cite{openai2024gpt4technicalreport}, or open-weight models, such as LLaMa~\cite{llama3} and DeepSeek~\cite{deepseekai2025deepseekr1incentivizingreasoningcapability} are inevitably prone to production of hallucinations or unsupported facts in their generated output~\cite{xiao-wang-2021-hallucination,dziri-etal-2022-origin,xu2024hallucinations}. This phenomenon poses a crucial obstacle for the real-world deployment of LLMs, particularly in safety-critical domains such as medicine~\cite{BenAbacha-BMC-2019,He2025medicine}. A single hallucinated word can substantially alter the overall meaning of the generation, potentially causing harm to end-users. Consequently, hallucination detection has become a critical challenge in the development and application of LLMs~\cite{huang2023hallucinations}.

\begin{figure*}
\centering
\includegraphics[width=\textwidth]{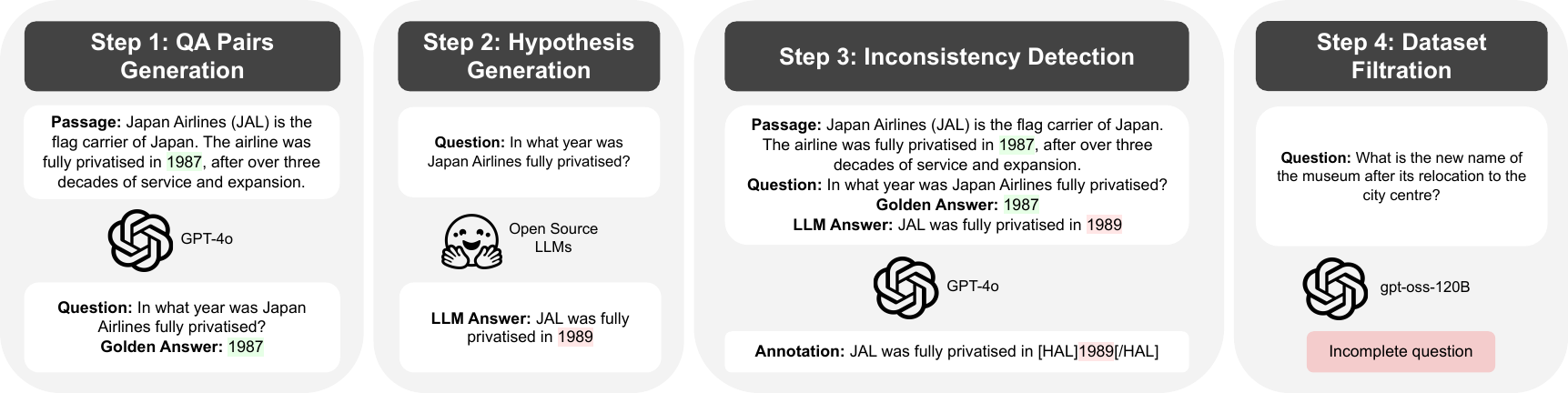}
\caption{PsiloQA generation pipeline, where a dataset is built in fourth steps. \textbf{Step 1:} Generation of multilingual question-answer pairs using the GPT-4o and randomly retrieved passages from Wikipedia articles.
\textbf{Step 2:} Generation of an answer to the question without the supporting passages from Wikipedia. By using only internal knowledge without external sources of information, LLMs cannot easily answer hard factual questions. \textbf{Step 3:} Span-level inconsistency detection between the golden answer generated by GPT-4o and the LLM hypothesis. \textbf{Step 4:} Filtration of incomplete or subjective questions and cases when LLM refuses to answer.}
\label{fig:psiloqa-generation}
\end{figure*}

Hallucination detection is typically categorized into three standard tasks: \textit{sequence-level}, \textit{span-level}, and \textit{entity-level}. Sequence-level detection focuses on identifying entire generations that contain some factual inconsistencies. In contrast, span-level and entity-level detection addresses the more challenging task of precisely highlighting factually misaligned spans or individual entities within the generated text.

Several approaches have been proposed for the detection of hallucination of LLMs. Uncertainty quantification (UQ) emerging as one of the most prominent research directions~\cite{gal2016dropout,shelmanov-etal-2021-certain,baan2023uncertainty,geng2023survey,vashurin2025Benchmarking,jellybell}. Recently, numerous UQ methods have been developed specifically for LLMs~\cite{kuhn2023semantic, lin2023generating}. However, most of these methods focus on sequence-level verification~\cite{farquhar2024detecting}, while only a few methods operate at the token or span level~\cite{zhang2023enhancing, fadeeva2024factchecking}. While \citet{rykov-etal-2025-smurfcat} proposed combining uncertainty estimation methods with a fine-tuned LLM, Qwen2.5-7B-Instruct, using a weighted averaging approach, where the weights were optimized separately for each language. 

Although UQ is a rapidly growing area of research, even the most advanced methods still face significant limitations. For instance, sampling-based methods~\cite{duan-etal-2024-shifting} require substantial computational overhead and operate only on the sequence level. Information-based methods~\cite{fomicheva-etal-2020-unsupervised,fadeeva2024factchecking} demonstrate strong performance in span-level tasks but still fail to detect some hallucinations and remain far from ideal performance. 

Another set of approaches focuses on fact-checking techniques based on external knowledge sources~\cite{niu-etal-2024-ragtruth} or auxiliary LLMs~\cite{fava}. While these methods achieve high performance, they require substantial computational overhead. These approaches first extract atomic claims from the generated response and compare them to a retrieved context using an auxiliary LLM~\cite{min-etal-2023-factscore}. This process produces a verification score, indicating the degree to which the extracted claims supported with the retrieved evidence. Moreover, the performance of such systems is heavily dependent on the quality of both the retrieved context and the auxiliary LLM, which typically requires fine-tuning for better performance~\cite{fava}.

To evaluate the quality of both systems, including those based on uncertainty quantification and external knowledge, we require a dataset with annotated hallucinations.
High-quality, fine-grained annotations, particularly at the span level, are labor-intensive, requiring expert human annotators~\cite{vazquez-etal-2025-mu-shroom} or costly automated pipelines~\cite{min-etal-2023-factscore}. Span-level annotation, though more practical for pinpointing unsupported text, introduces additional complexity compared to sequence-level verification. Multilingual contexts increase these challenges due to linguistic diversity, limited non-English data availability, and the need for language-specific pipeline adjustments~\cite{vashurin2025Benchmarking}.

To address these limitations, we introduce a novel methodology for automatically generating multilingual data with fine-grained hallucination annotations, which includes
(i) synthetically creating question-answering pairs from Wikipedia article summaries, (ii) generating hypotheses using LLMs in a zero-context setting to produce both hallucinated and accurate answers, (iii) automating span-level inconsistency annotation by comparing responses to context and ground truth via an advanced LLM, and (iv) automated filtering step through both rule-based and prompt-based methods. Further details of this pipeline are outlined in Section~\ref{sec:psiloqa}.

The contributions of this work could be summarized as follows:
\begin{itemize}
  \itemsep 0ex 
  \topsep 0ex 
  \item We propose an automated and scalable pipeline for generating and annotating synthetic data.
  \item We introduce a large multilingual dataset with high-quality and fine-grained span-level hallucination annotations for numerous open-weighted and proprietary LLMs.
  \item We conduct comprehensive empirical evaluations of various state-of-the-art hallucination detection methods of different types across 14 languages.
\end{itemize}

\begin{table*}
% \scriptsize
\centering
\resizebox{\textwidth}{!}{%
\begin{tabular}{lcccccccccc}
\toprule
\textbf{Dataset} & \textbf{Domain} & \textbf{Annotation} & \textbf{Generation} & \textbf{Lang} & \textbf{\# LLMs} & \textbf{\# Train} & \textbf{\# Val} & \textbf{\# Test} & \textbf{Licence}\\
\midrule
Mu-SHROOM~\cite{vazquez-etal-2025-mu-shroom} 
& General & Manual & Natural & Mult & 38 & 3,351$^*$ & 499 & 1,902 & CC-BY-4.0\\
HalluEntity~\cite{halluentity} 
& Biography & Manual & Natural & En & 1 & -- & -- & 157 & MIT\\
RAGTruth\textsubscript{QA}~\cite{niu-etal-2024-ragtruth} 
& General & Manual & Natural & En & 6 & 5,034 & -- & 900 & MIT\\
FAVA-Bench~\cite{fava} 
& General & Auto & Synthetic & En & 3 & -- & -- & 902 & CC-BY-4.0\\
\midrule
PsiloQA (ours) 
& General & Auto & Natural & Mult & 24 & 63,792 & 3,355 & 2,897 & CC-BY-4.0\\
\bottomrule
\end{tabular}
}
\caption{Comparative overview of span-level hallucination detection datasets. The Mu-SHROOM dataset has an unlabeled training set ($^*$) comprising 4 languages (en, es, fr, zh). The \textbf{Generation} column distinguishes whether LLM answers were generated with intentional error insertion (synthetic) or used as-is (natural).}
\label{table:datasets}
\end{table*}

\section{Related Work}
\subsection{Hallucination Detection Datasets}
Most hallucination detection benchmarks operate at the sentence or paragraph level, such as TruthfulQA~\cite{truthfulqa}, ANAH~\cite{ji-etal-2024-anah, anah_v2} and HaluEval~\cite{halueval}. These benchmarks categorize each generated response as either hallucinated or correct. However, instance-level detection is unable to identify specific hallucinated content, which is essential for correcting misinformation. This limitation is especially concerning in long-form text where a single response may include both supported and unsupported information, rendering binary quality assessments insufficient~\cite{min-etal-2023-factscore}.

To tackle these issues, recent studies have improved benchmarks for more detailed hallucination detection. For instance, ~\citet{min-etal-2023-factscore} introduced FActScore, a dataset focusing on fine-grained hallucination detection in Wikipedia bibliographies. It aids in assessing hallucination detection techniques that utilize external knowledge and detailed fact-level annotations.

Similarly, HalluEntity~\cite{halluentity} includes biographies created by ChatGPT, with each entry comprising a name, a ChatGPT-generated biography, and a list of atomic facts marked as True or False, aligning with the relevant entity in the language model's output.

RAGTruth~\cite{niu-etal-2024-ragtruth} is a large-scale benchmark with nearly 18,000 human-annotated examples designed for retrieval-augmented generation (RAG) tasks. It provides fine-grained word-level annotations marking hallucinated spans that contradict reference documents, covering question answering, data-to-text, and summarization tasks.

FAVA (FavaBench)~\cite{fava} is a dataset containing fine-grained hallucination annotations with external knowledge support, used to evaluate hallucination detection and editing. It includes diverse generation tasks and has been shown effective in fine-grained hallucination detection research.

The multilingual Mu-SHROOM dataset has been suggested in the shared task on Multilingual Hallucinations and Related Observable Overgeneration Mistakes~\cite{vazquez-etal-2025-mu-shroom}. In particular, the Mu-SHROOM task aims to detect hallucination spans in the outputs of instruction-tuned LLMs in multilingual context models for Arabic, Basque, Catalan, Chinese, Czech, English, Farsi, Finnish, French, German, Hindi, Italian, Spanish, and Swedish.\footnote{\url{https://helsinki-nlp.github.io/shroom}} The datasets with their respective annotation levels and splits are outlined in Table~\ref{table:datasets}.

\subsection{Hallucination Detection Methods}
Most hallucination detection methods operate at the sentence-level. For instance, recent works propose various sampling-based UQ methods that measure the consistency of multiple sampled generations~\cite{kuhn2023semantic,lin2023generating,duan-etal-2024-shifting,zhang-etal-2024-luq}. On the contrary, reflexive methods aims to assess an LLM's confidence in its generation by directly prompting it for self-evaluation~\cite{kadavath2022language,tian-etal-2023-just}. Supervised methods are applicable at both sequence and token levels, but they often require model-specific training due to differences in number of hidden features and attention heads~\cite{azaria-mitchell-2023-internal,vazhentsev2024unconditional,chuang-etal-2024-lookback,ch-wang-etal-2024-androids,vazhentsev-etal-2025-token}. 

Token probability and entropy~\cite{fomicheva-etal-2020-unsupervised} are trivial baselines for span-level detection, which utilize the distribution of token probabilities. \citet{fadeeva2024factchecking} propose to analyze the consistency of the top most probable token candidates by leveraging the natural language inference (NLI) model using the Claim Conditioned Probability (CCP) method. \citet{zhang2023enhancing} propose to model the conditional dependencies between the generated tokens by reweighing token uncertainty scores, leveraging uncertainty of the previous tokens and attention weights after max-pooling.

Several methods leverage external knowledge to evaluate the factuality of the generations. Among the most well-known is FActScore~\cite{min-etal-2023-factscore}, which extracts atomic facts from the model's response and compares them to a retrieved context using an additional LLM. This fact-checking process generates a score that indicates whether the claims are supported by the retrieved context. 

\citet{niu-etal-2024-ragtruth} introduce RAGTruth, a pipeline for the detection of word-level hallucinations in retrieval-augmented generation (RAG) systems. This work introduces a dataset and benchmark to evaluate the factuality of LLM responses for various tasks, such as summarization, question-answering, and others. Moreover, this framework could be easily adapted for the fact-checking task. The hallucinations annotations in the presented data were created by human annotators.

Furthermore, the task of hallucination detection can naturally be extended to hallucination editing. For example, the FAVA~\cite{fava} model is specifically trained for word-level hallucination detection and editing tasks according to the introduced hallucination taxonomy. To collect training data, the authors asked LLMs to insert errors from the introduced taxonomy into the responses.

Despite their advantages, both RAGTruth and FAVA are limited in their applicability, as they are designed only for English-language tasks and require human annotations.

\section{PsiloQA: A Synthetic Span-Level Hallucination Dataset}
\label{sec:psiloqa}
\subsection{Dataset Generation Process}

Figure~\ref{fig:psiloqa-generation} illustrates the dataset generation pipeline. Our objectives in developing the PsiloQA generation process include: (i) utilizing real LLM hallucinations rather than artificially inserted errors; (ii) ensuring the process is cost-effective, quick, and scalable; (iii) encompassing multiple languages and domains.

The initial stage of the PsiloQA pipeline involves generating context-based question–answer pairs. Given the scarcity of such multilingual data in the general domain, we constructed a multilingual context-based QA dataset from scratch. This is achieved by utilizing passages from Wikipedia to source diverse, multilingual data. The passages, along with a specific prompt, are submitted to GPT-4o to generate QA pairs. To achieve varying question complexity, 3 different question-answer pairs are produced with varying levels of complexity, as demonstrated in Figure~\ref{fig:qa-generation-prompt}.

To generate answers that contain hallucinations, we asked various LLMs to respond to previously generated questions without referring to Wikipedia for support. When relying solely on internal knowledge, LLMs often produce hallucinations in their responses to factual questions. The same models employed in Mu-SHROOM~\cite{vazquez-etal-2025-mu-shroom} were used to generate these inaccurate answers.

To catch hallucinations, we prompt GPT-4o to review the passage and question, comparing the golden answer with the LLM hypothesis. Any discrepancies are marked with \texttt{[HAL]} tags. Otherwise, the LLM's response is copied unchanged when no inconsistencies are found. The prompt for detecting inconsistencies is illustrated in Figure~\ref{fig:inconsistency-detection-prompt}. For span-level annotation, we followed the RAGTruth pipeline~\cite{niu-etal-2024-ragtruth} and asked GPT-4o to annotate spans at the word level, encouraging precise labeling and discouraging overgeneralization (e.g., marking the entire answer as a hallucination).

Additionally, we conducted several automatic filtering steps, including both rule-based and prompt-based. The rule-based filter removes samples where \texttt{[HAL]} tags did not properly match, cases where the annotator model generated empty spans, and cases where the annotated answer is not consistent with the initial LLM's answer after removing all \texttt{[HAL]} tags. The prompt-based filter removes subjective questions, incomplete questions, and answers where LLM refuses to answer. Subjective questions are non-factual and thus do not require context for answering. Consequently, it is challenging to recognize any inconsistencies when comparing the answers to subjective questions with the original contexts. Incomplete questions are artifacts of the QA pairs generation process, as they have a reference to the original context, or use pronouns with no clear antecedent. The absence of an explicit subject in a question makes it unanswerable without context. Finally, cases when LLM refuses to answer also introduce difficulties in identifying inconsistent segments, as they do not constitute responses to factual questions. Prompt-based filtering was performed using \texttt{gpt-oss-120B}\footnote{\url{https://hf.co/openai/gpt-oss-120b}} model. Instructions are shown in Appendix~\ref{app:filtering-prompts}. In total, 6,575 samples were filtered.

Consequently, PsiloQA fulfills all our previously outlined criteria. It utilizes LLM for question answering, which results in samples containing genuine hallucinations. Furthermore, using GPT-4o for annotation makes the generation process scalable, more cost-effective, and faster compared to human annotation. Wikipedia is used as a dependable and varied source of multilingual seeds.

After all filtration steps, the training set of PsiloQA consists of 63,792 samples. For each language and LLM checkpoint combination, we select 100 random samples for benchmarking. The PsiloQA testing split contains 2,897 samples.

\subsection{Dataset Statistics}
Figure~\ref{fig:lang-stats} illustrates the distribution of samples by language in the PsiloQA dataset. English is the most prevalent, with nearly 23,000 samples. Hindi, Finnish, Catalan, Chinese, Swedish, and Czech each range between 5,000 and 7,000 samples. The dataset contains roughly 3,700 Farsi samples, and approximately 2,000 to 2,500 samples per language for Spanish, Euskara, French, Italian, and Arabic. German appears as the least represented language, with about 1,500 samples. Figure~\ref{fig:model-stats} displays the statistics by LLMs.

Figure~\ref{fig:span-count} illustrates the number of hallucination spans present in each sample. There are 14,000 samples with no hallucinations and 50,000 samples containing just one span of hallucination. Other samples have 2 or more spans, with a maximum of 10 in rare cases. Figure~\ref{fig:span-length} depicts the word distribution of span lengths. PsiloQA spans are relatively short, with 50,000 spans including fewer than 5 words.

Also, we analyzed the distribution of predicted domains in the PsiloQA dataset. Each passage was assigned a single domain using zero-shot classification with the \texttt{bart-large-mnli}\footnote{\url{https://hf.co/facebook/bart-large-mnli}} model across $34$ candidate domains. Geography and Sports are the most prevalent, with roughly $25$–$30\%$ of samples each. Overall, the dataset exhibits a diverse distribution, with some highly represented domains and a long tail of less frequent categories.

\subsection{Dataset Production Cost}
The estimated cost of GPT-4o labeling is \$535, based on token generation pricing where \$4 is charged for every 1M input tokens and \$16 for every 1M tokens. To compare this with RAGTruth, the only span-level hallucination dataset with a labeled training set, we estimated its cost. To annotate RAGTruth, annotators proficient in English and holding a bachelor’s degree in English, Communications, or relevant fields were employed to ensure accuracy and reliability. They were recruited from a professional vendor and compensated at \$25 per hour per individual. Each response was labeled by two annotators, achieving a 91.8\% consistency rate at the response level and 78.8\% at the span level. 
The total cost of labeling is not specified in the paper, but our rough estimation suggests the span-level labeling of RAGTruth\textsubscript{QA} was approximately \$3,000.

\subsection{Manual Dataset Analysis}
To validate the quality of our automatic annotation pipeline, we conducted a manual verification study on 100 randomly selected samples from the English PsiloQA test split. Three annotators with MS degrees in relevant fields were tasked with identifying hallucination spans within these samples, using general instruction presented in Figure~\ref{fig:inconsistency-detection-prompt}.

Following our dual-level evaluation approach (detailed in Section~\ref{sec:manual-analysis}), we assessed annotation quality using two metrics: average precision (AP) and intersection over union (IoU). For inter-annotator agreement, we computed the mean of all pairwise comparisons between annotators, yielding an AP of 80.1\% and IoU of 76.8\%, demonstrating substantial consensus among human annotators. To compare human labels against GPT-4o's automatic predictions, we first aggregated the three manual annotations: for IoU, we computed the character-level union, while for AP, we calculated the mean score for each character position. The aggregated reference showed strong alignment with GPT-4o's predictions, achieving an AP of 84.3\% and IoU of 71.0\%.

The result indicates that GPT-4o is a reliable annotator of span-level hallucination given ground-truth context. The chosen sample size yields a worst-case margin of error of approximately 9.8\% at the 95\% confidence level~\cite{klie-etal-2024-efficient}, providing reasonable confidence of the pipeline’s overall adequacy. To further validate annotation quality, we present cross-lingual transfer results in Section~\ref{sec:transfer} using the Mu-SHROOM~\cite{vazquez-etal-2025-mu-shroom} dataset.

\section{Experimental Setup}
\label{sec:manual-analysis}
In our experiments, we pursue the following objectives: (i) to evaluate the performance of uncertainty quantification (UQ) baselines, large language models (LLMs), and state-of-the-art methods on the PsiloQA dataset; (ii) to demonstrate the transferability of knowledge from models trained on PsiloQA and RAGTruth to a range of downstream benchmarks;
(iii) to assess the impact of multilingual training in PsiloQA by comparing two configurations of mmBERT~\cite{mmbert}: one trained on the full multilingual PsiloQA dataset and the other trained separately on each individual language subset.

\subsection{Datasets}
In addition to PsiloQA (described in Section~\ref{sec:psiloqa}), we employ four different QA-based benchmarks to evaluate several methods for identifying span-level hallucinations, with the dataset specifics outlined in Table~\ref{table:datasets}.

\noindent\textbf{Mu-SHROOM}: a multilingual benchmark for 14 languages. The dataset contains 3,351 unlabeled samples in four languages: English, Spanish, Chinese, French. The test set contains 1,902 samples (Basque, Catalan, Czech and Farsi containing around 100 items, and other languages containing around 150 items).

\noindent\textbf{FAVA-Bench}: An English-language, human-annotated benchmark designed to identify and correct various types of hallucinations according to the FAVA taxonomy. Due to the inability of most models to detect errors using this taxonomy, the original benchmark's focus was limited to span-level hallucination detection.

\noindent\textbf{HalluEntity}: A benchmark in English for detecting entity-level hallucinations, comprising 157 human-annotated samples. Annotations were gathered by having ChatGPT generate biographies of various well-known individuals.

\noindent\textbf{RAGTruth\textsubscript{QA}}: An English-language, human-annotated benchmark for detecting hallucinations, consisting of 900 samples with questions sourced from the MS MARCO dataset. For generating responses, six different models were utilized: \texttt{GPT-3.5-turbo-0613}, \texttt{GPT-4-0613}, \texttt{Llama-2-7B-chat}, \texttt{Llama-2-13B-chat}, \texttt{Llama-2-70B-chat} and \texttt{Mistral-7B-Instruct}.

\subsection{Metrics}
Due to the complex nature of hallucination detection, we employ a dual-level evaluation approach combining span-level and character-level assessment. As with Mu-SHROOM, we selected the intersection over union (IoU) metric to evaluate span-level hallucination detection. Additionally, we use average precision (AP) for ranking-based evaluation on character-level.

First, the span-level annotation is converted into a set of binary labels for each character. Next, the IoU is calculated as follows:
\begin{align}
    \mathrm{IoU} &= {\left|\hat{C}_\mathrm{bin} \cap C_\mathrm{bin}\right|} ~/~ {\left|\hat{C}_\mathrm{bin} \cup C_\mathrm{bin}\right|},
\end{align}
\noindent where $C_\mathrm{bin}$ is the set of binarized character-level annotations, and $\hat{C}_\mathrm{bin}$ is the set of characters that the model predicts as hallucinated.

Average precision provides a threshold-independent evaluation by computing the area under the precision-recall curve. In practice, it is calculated as:
\begin{align}
    \mathrm{AP} &= {\sum\limits_n p(n) \Delta r(n)}
\end{align}
\noindent where $p(n)$ is the precision at cut-off $n$ in the ranked list and $\Delta r(n)$ is the change in recall between items $n-1$ and $n$. This metric is particularly valuable for hallucination detection as it handles imbalanced datasets effectively and provides robust evaluation across different prediction confidence distributions, making it suitable for both in-domain evaluation and cross-dataset transfer scenarios.

\begin{table*}[htb!] 
\centering
\resizebox{\linewidth}{!}{ 
\small
\begin{tabular}{llcccccccccccccccc} 
\toprule
\textbf{Method} & \textbf{Mode} & \textbf{Metrics} & \textbf{ar} & \textbf{ca} & \textbf{cs} & \textbf{de} & \textbf{en} & \textbf{es} & \textbf{eu} & \textbf{fa} & \textbf{fi} & \textbf{fr} & \textbf{hi} & \textbf{it} & \textbf{sv} & \textbf{zh} \\ 
\midrule
\multicolumn{17}{l}{\textit{Uncertainty Quantification}} \\ 
\hspace{0.2cm}\multirow{2}{*}{MSP} & \multirow{2}{*}{--} & AP & 43.38 & 39.41 & 40.12 & 30.86 & 59.38 & 52.04 & 50.76 & 41.08 & 65.95 & 49.84 & 56.75 & 47.71 & 45.39 & 49.18 \\ 
 & & \cellcolor{gray!20}IoU & \cellcolor{gray!20}35.70 & \cellcolor{gray!20}28.36 & \cellcolor{gray!20}33.68 & \cellcolor{gray!20}30.03 & \cellcolor{gray!20}45.69 & \cellcolor{gray!20}33.72 & \cellcolor{gray!20}33.04 & \cellcolor{gray!20}22.13 & \cellcolor{gray!20}53.13 & \cellcolor{gray!20}37.67 & \cellcolor{gray!20}43.45 & \cellcolor{gray!20}31.61 & \cellcolor{gray!20}26.96 & \cellcolor{gray!20}28.42 \\ 
\hspace{0.2cm}\multirow{2}{*}{CCP} & \multirow{2}{*}{--} & AP & 48.90 & 41.17 & 40.98 & 31.18 & 62.52 & 52.55 & 51.63 & 44.87 & 66.75 & 50.43 & 59.62 & 49.75 & 45.30 & 52.67 \\ 
 & & \cellcolor{gray!20}IoU & \cellcolor{gray!20}35.70 & \cellcolor{gray!20}28.37 & \cellcolor{gray!20}33.68 & \cellcolor{gray!20}33.25 & \cellcolor{gray!20}45.69 & \cellcolor{gray!20}33.72 & \cellcolor{gray!20}33.04 & \cellcolor{gray!20}22.13 & \cellcolor{gray!20}53.13 & \cellcolor{gray!20}37.67 & \cellcolor{gray!20}43.45 & \cellcolor{gray!20}32.20 & \cellcolor{gray!20}26.96 & \cellcolor{gray!20}27.39 \\ 
\hspace{0.2cm}\multirow{2}{*}{Focus} & \multirow{2}{*}{--} & AP & 49.87 & 39.88 & 43.86 & 32.71 & 63.61 & 61.72 & 52.84 & 47.12 & 68.90 & 53.65 & 60.09 & 48.07 & 56.20 & 53.05 \\ 
 & & \cellcolor{gray!20}IoU & \cellcolor{gray!20}36.93 & \cellcolor{gray!20}28.37 & \cellcolor{gray!20}33.68 & \cellcolor{gray!20}32.05 & \cellcolor{gray!20}45.69 & \cellcolor{gray!20}42.24 & \cellcolor{gray!20}34.65 & \cellcolor{gray!20}29.94 & \cellcolor{gray!20}53.13 & \cellcolor{gray!20}39.26 & \cellcolor{gray!20}43.45 & \cellcolor{gray!20}32.20 & \cellcolor{gray!20}36.15 & \cellcolor{gray!20}27.83 \\ 
\midrule
\multicolumn{17}{l}{\textit{Encoder Models}} \\ 
\hspace{0.2cm}\multirow{2}{*}{lettuce-detect-base} & \multirow{2}{*}{--} & AP & 46.23 & 57.71 & 32.53 & 32.15 & 54.21 & 51.27 & 30.78 & 32.45 & 57.43 & 37.51 & 33.17 & 35.36 & 48.97 & 31.01 \\ 
 & & \cellcolor{gray!20}IoU & \cellcolor{gray!20}37.81 & \cellcolor{gray!20}44.37 & \cellcolor{gray!20}30.08 & \cellcolor{gray!20}30.31 & \cellcolor{gray!20}43.28 & \cellcolor{gray!20}40.08 & \cellcolor{gray!20}33.35 & \cellcolor{gray!20}32.45 & \cellcolor{gray!20}56.44 & \cellcolor{gray!20}35.60 & \cellcolor{gray!20}16.95 & \cellcolor{gray!20}34.97 & \cellcolor{gray!20}49.11 & \cellcolor{gray!20}35.94 \\ 
\hspace{0.2cm}\multirow{2}{*}{ModernBERT-base} & \multirow{2}{*}{SFT} & AP & \underline{60.37} & \underline{75.48} & 53.46 & 44.77 & \underline{81.63} & \underline{81.71} & 58.72 & 53.84 & 66.87 & 68.48 & 71.94 & \underline{72.00} & \underline{79.94} & 66.84 \\ 
 & & \cellcolor{gray!20}IoU & \cellcolor{gray!20}\underline{55.27} & \cellcolor{gray!20}\underline{65.70} & \cellcolor{gray!20}44.73 & \cellcolor{gray!20}\underline{46.27} & \cellcolor{gray!20}\underline{68.23} & \cellcolor{gray!20}\underline{61.69} & \cellcolor{gray!20}\textbf{50.43} & \cellcolor{gray!20}\underline{68.63} & \cellcolor{gray!20}\underline{64.68} & \cellcolor{gray!20}\underline{53.90} & \cellcolor{gray!20}\underline{54.15} & \cellcolor{gray!20}\underline{62.75} & \cellcolor{gray!20}\textbf{67.09} & \cellcolor{gray!20}\underline{56.95} \\ 
\hspace{0.2cm}\multirow{2}{*}{mmBERT-base} & \multirow{2}{*}{SFT} & AP & \textbf{70.71} & \textbf{77.22} & \textbf{67.62} & \underline{61.40} & \textbf{84.88} & \textbf{84.84} & \underline{65.30} & \textbf{75.24} & \textbf{75.85} & \textbf{73.52} & \textbf{78.33} & \textbf{73.81} & \textbf{84.04} & \underline{73.79} \\ 
 & & \cellcolor{gray!20}IoU & \cellcolor{gray!20}\textbf{58.10} & \cellcolor{gray!20}\textbf{67.01} & \cellcolor{gray!20}\textbf{48.81} & \cellcolor{gray!20}\textbf{54.97} & \cellcolor{gray!20}\textbf{70.67} & \cellcolor{gray!20}\textbf{66.18} & \cellcolor{gray!20}\underline{50.27} & \cellcolor{gray!20}\textbf{76.61} & \cellcolor{gray!20}\textbf{68.16} & \cellcolor{gray!20}\textbf{56.38} & \cellcolor{gray!20}\textbf{61.19} & \cellcolor{gray!20}\textbf{66.57} & \cellcolor{gray!20}\underline{66.24} & \cellcolor{gray!20}\textbf{61.58} \\ 
\midrule
\multicolumn{17}{l}{\textit{Language Models}} \\ 
\hspace{0.2cm}\multirow{2}{*}{FActScore (GPT-4o)} & \multirow{2}{*}{--} & AP & 53.62 & 45.24 & 58.65 & 43.32 & 62.38 & 51.75 & \textbf{66.82} & \underline{70.04} & \underline{74.49} & \underline{71.12} & 50.35 & 69.81 & 69.68 & 58.68 \\ 
 & & \cellcolor{gray!20}IoU & \cellcolor{gray!20}20.75 & \cellcolor{gray!20}28.99 & \cellcolor{gray!20}10.44 & \cellcolor{gray!20}26.68 & \cellcolor{gray!20}25.84 & \cellcolor{gray!20}28.54 & \cellcolor{gray!20}19.68 & \cellcolor{gray!20}26.62 & \cellcolor{gray!20}28.16 & \cellcolor{gray!20}10.21 & \cellcolor{gray!20}21.03 & \cellcolor{gray!20}43.92 & \cellcolor{gray!20}19.25 & \cellcolor{gray!20}25.18 \\ 
\hspace{0.2cm}\multirow{2}{*}{Qwen2.5-32B-it} & \multirow{2}{*}{3-shot} & AP & 54.52 & 67.66 & \underline{61.18} & \textbf{70.50} & 63.17 & 54.69 & 57.10 & 59.68 & 72.42 & 67.41 & \underline{76.14} & 57.65 & 64.43 & \textbf{77.20} \\ 
 & & \cellcolor{gray!20}IoU & \cellcolor{gray!20}35.54 & \cellcolor{gray!20}51.71 & \cellcolor{gray!20}\underline{46.83} & \cellcolor{gray!20}23.57 & \cellcolor{gray!20}39.98 & \cellcolor{gray!20}40.51 & \cellcolor{gray!20}36.52 & \cellcolor{gray!20}19.18 & \cellcolor{gray!20}34.69 & \cellcolor{gray!20}31.92 & \cellcolor{gray!20}44.56 & \cellcolor{gray!20}37.95 & \cellcolor{gray!20}50.89 & \cellcolor{gray!20}42.77 \\ 
\bottomrule
\end{tabular} 
}
\caption{Performance comparison of span-level hallucination detection methods on the PsiloQA test set across 14 languages. Encoder models were supervised fine-tuned (SFT) on the complete PsiloQA train set, while Qwen2.5-32B-it used 3-shot prompting.}
\label{tab:results}
\end{table*}

\subsection{Baselines}
\subsubsection{Uncertainty Quantification}
Most uncertainty quantification methods either operate at the sequence-level or require model-specific training. Therefore, for a given generated text $\tilde{y}$ of a length $N$, for each token $t_i\in\tilde{y}$, $i=1\dots N$, we compute three uncertainty quantification methods, which are designed for token-level tasks.  The set of baselines includes Maximum Token Probability (MaxProb; \citet{fomicheva-etal-2020-unsupervised}), Claim Conditioned Probability (CCP; \citet{fadeeva2024factchecking}), and Focus~\cite{zhang2023enhancing}. To compute the IoU metrics, we employ language-specific threshold calibration on the validation set.

\subsubsection{Encoder Models}
We evaluate several encoder-based transformer models fine-tuned for token-level hallucination detection. These models process context-question-answer triples to identify unsupported claims at the token level.
LettuceDetect models~\cite{lettucedetect}, built on ModernBERT~\cite{modernbert}, were trained on the RAGTruth dataset and offer extended context processing capabilities (8,192 tokens) through a local-global attention mechanism~\cite{gemma}.
However, they were pre-trained primarily on English data from various sources like web documents, code, and scientific literature. This implies that the models are not directly suitable for use with other languages, and their effectiveness may be further reduced for low-resource languages, even though the tokenizers might include subtokens pertaining to non-English languages.

To address multilingual requirements, we fine-tuned \texttt{mmBERT-base}\footnote{\url{https://hf.co/jhu-clsp/mmBERT-base}}, Modern Multilingual Encoder (307M parameters) that extends ModernBERT with native support for multiple languages. We also fine-tuned \texttt{ModernBERT-base}\footnote{\url{https://hf.co/answerdotai/ModernBERT-base}} on the complete PsiloQA dataset for comparison. All models were fine-tuned using identical hyperparameters as those in LettuceDetect, detailed in Table~\ref{tab:encoderhyper} (Appendix~\ref{app:encoderhyper}).

\subsubsection{Language Models}
We also use two LLM-based approaches.
FActScore~\cite{min-etal-2023-factscore} decomposes model responses into atomic facts and verifies each against the provided context using an LLM (GPT-4o in our implementation). To adapt this sentence-level method for token-level annotation, we compute token-level hallucination scores based on their frequency in unsupported claims, applying a threshold of 0.5 for binary classification. Tokens appearing in all claims or present in the original input are excluded from hallucination marking. Few-shot prompting with \texttt{Qwen2.5-32B-Instruct}\footnote{\url{https://hf.co/Qwen/Qwen2.5-32B-Instruct}} using 3-shot learning with examples randomly selected from the validation set. The prompting template is detailed in Figure~\ref{fig:baseline-prompt}.

\begin{table*}[htb!]
\centering
\resizebox{\linewidth}{!}{
\begin{tabular}{ll cccccccccccccc} 
\toprule
\textbf{Strategy} & \textbf{Metrics} & \textbf{ar} & \textbf{ca} & \textbf{cs} & \textbf{de} & \textbf{en} & \textbf{es} & \textbf{eu} & \textbf{fa} & \textbf{fi} & \textbf{fr} & \textbf{hi} & \textbf{it} & \textbf{sv} & \textbf{zh} \\ 
\midrule
\multicolumn{16}{l}{\textit{PsiloQA}} \\ 
\hspace{0.2cm}\multirow{2}{*}{Per language} & \cellcolor{gray!20}IoU & \cellcolor{gray!20}45.7 & \cellcolor{gray!20}59.88 & \cellcolor{gray!20}45.64 & \cellcolor{gray!20}39.68 & \cellcolor{gray!20}\textbf{72.05} & \cellcolor{gray!20}55.05 & \cellcolor{gray!20}42.33 & \cellcolor{gray!20}69.37 & \cellcolor{gray!20}61.98 & \cellcolor{gray!20}49.2 & \cellcolor{gray!20}52.11 & \cellcolor{gray!20}59.27 & \cellcolor{gray!20}62.74 & \cellcolor{gray!20}51.62 \\ 
 & AP & 57.1 & 70.79 & 53.71 & 58.29 & 82.82 & 67.71 & 49.81 & 65.54 & 65.1 & 62.57 & 71.3 & 69.88 & 83.69 & 69.73 \\ 
\hspace{0.2cm}\multirow{2}{*}{Multilingual} & \cellcolor{gray!20}IoU & \cellcolor{gray!20}\textbf{58.1} & \cellcolor{gray!20}\textbf{67.01} & \cellcolor{gray!20}\textbf{48.81} & \cellcolor{gray!20}\textbf{54.97} & \cellcolor{gray!20}70.67 & \cellcolor{gray!20}\textbf{66.18} & \cellcolor{gray!20}\textbf{50.27} & \cellcolor{gray!20}\textbf{76.61} & \cellcolor{gray!20}\textbf{68.16} & \cellcolor{gray!20}\textbf{56.38} & \cellcolor{gray!20}\textbf{61.19} & \cellcolor{gray!20}\textbf{66.57} & \cellcolor{gray!20}\textbf{66.24} & \cellcolor{gray!20}\textbf{61.58} \\ 
 & AP & \textbf{70.71} & \textbf{77.22} & \textbf{67.62} & \textbf{61.4} & \textbf{84.88} & \textbf{84.84} & \textbf{65.3} & \textbf{75.24} & \textbf{75.85} & \textbf{73.52} & \textbf{78.33} & \textbf{73.81} & \textbf{84.04} & \textbf{73.79} \\ 
\midrule
\multicolumn{16}{l}{\textit{Mu-SHROOM}} \\ 
\hspace{0.2cm}\multirow{2}{*}{Per language} & \cellcolor{gray!20}IoU & \cellcolor{gray!20}47.17 & \cellcolor{gray!20}52.4 & \cellcolor{gray!20}31.43 & \cellcolor{gray!20}38.42 & \cellcolor{gray!20}\textbf{58.51} & \cellcolor{gray!20}31.27 & \cellcolor{gray!20}38.11 & \cellcolor{gray!20}49.62 & \cellcolor{gray!20}\textbf{61.69} & \cellcolor{gray!20}53.53 & \cellcolor{gray!20}62.91 & \cellcolor{gray!20}61.27 & \cellcolor{gray!20}33.58 & \cellcolor{gray!20}31.05 \\ 
 & AP & 67.18 & 64.33 & 51.05 & 60.34 & 70.18 & 44.7 & 51.85 & 68.91 & \textbf{81.37} & 76.93 & 75.91 & 79.07 & 70.52 & \textbf{60.8} \\ 
\hspace{0.2cm}\multirow{2}{*}{Multilingual} & \cellcolor{gray!20}IoU & \cellcolor{gray!20}\textbf{65.87} & \cellcolor{gray!20}\textbf{65.12} & \cellcolor{gray!20}\textbf{42.14} & \cellcolor{gray!20}\textbf{64.13} & \cellcolor{gray!20}58 & \cellcolor{gray!20}\textbf{45.28} & \cellcolor{gray!20}\textbf{48.16} & \cellcolor{gray!20}\textbf{69.01} & \cellcolor{gray!20}51.69 & \cellcolor{gray!20}\textbf{59.49} & \cellcolor{gray!20}\textbf{71.69} & \cellcolor{gray!20}\textbf{72.6} & \cellcolor{gray!20}\textbf{49.45} & \cellcolor{gray!20}\textbf{38.19} \\ 
 & AP & \textbf{82.4} & \textbf{78.87} & \textbf{63.5} & \textbf{81.51} & \textbf{72.27} & \textbf{53.12} & \textbf{66.45} & \textbf{78.61} & 78.51 & \textbf{80.88} & \textbf{79.73} & \textbf{84.02} & \textbf{74.7} & 56.01 \\ 
\bottomrule
\end{tabular}  
}
\caption{Cross-lingual transfer results comparing two training strategies of mmBERT-base: language-specific models trained independently on each language subset of PsiloQA (per language) versus a single multilingual model trained on the complete PsiloQA dataset (multilingual). Both approaches are evaluated on test sets from PsiloQA and Mu-SHROOM datasets.}
\label{tab:cross_ling}
\end{table*}

\begin{table}[htb!]
\small
\centering
\resizebox{\linewidth}{!}{
\begin{tabular}{llccc} 
\toprule
\multirow{2}{*}{\textbf{Test}} & \multirow{2}{*}{\textbf{Metrics}} & \multicolumn{3}{c}{\textbf{Train}} \\
\cmidrule(lr){3-5}
 & & \textbf{RAGTruth\textsubscript{QA}} & \textbf{PsiloQA\textsubscript{en}} & \textbf{Both} \\ 
\midrule
\multirow{2}{*}{FAVA-Bench} & \cellcolor{gray!20}IoU & \cellcolor{gray!20}\underline{14.46} & \cellcolor{gray!20}14.29 & \cellcolor{gray!20}\textbf{14.88} \\ 
 & AP & \underline{18.55} & \textbf{23.10} & 17.36 \\ 
\midrule
\multirow{2}{*}{HalluEntity} & \cellcolor{gray!20}IoU & \cellcolor{gray!20}\underline{28.12} & \cellcolor{gray!20}\textbf{30.80} & \cellcolor{gray!20}25.53 \\ 
 & AP & 40.94 & \underline{56.33} & \textbf{63.37} \\ 
\midrule
\multirow{2}{*}{Mu-SHROOM\textsubscript{en}} & \cellcolor{gray!20}IoU & \cellcolor{gray!20}40.27 & \cellcolor{gray!20}\textbf{58.51} & \cellcolor{gray!20}\underline{55.90} \\ 
 & AP & 46.45 & \textbf{70.18} & \underline{67.31} \\ 
\bottomrule
\end{tabular}
}
\caption{Generalization performance of mmBERT-base across different hallucination detection benchmarks. Models were fine-tuned on RAGTruth\textsubscript{QA}, PsiloQA\textsubscript{en}, or both datasets, and evaluated on FAVA-Bench, HalluEntity, and Mu-SHROOM\textsubscript{en} test sets.}
\label{tab:ragtruth_vs_psilo}
\end{table}

\section{Results}

\subsection{Performance on PsiloQA}
Table~\ref{tab:results} presents the performance of span-level hallucination detection methods on PsiloQA across 14 languages.

\textit{Uncertainty Quantification} methods show moderate performance, with Focus consistently outperforming MSP and CCP across both metrics, achieving the highest AP scores (e.g., 68.90 for Finnish, 63.61 for English) among UQ approaches. However, IoU scores remain relatively low across all languages, indicating limited precision in span-level detection.

\textit{Encoder models} demonstrate superior performance, with a clear hierarchy emerging. The pre-trained LettuceDetect model shows mixed results, performing reasonably on some languages. Fine-tuned models significantly outperform the pre-trained baseline: ModernBERT achieves strong results, while mmBERT obtains the best overall performance, achieving the highest scores in 12 of 14 languages for both metrics. This superiority highlights the importance of multilingual pre-training for cross-lingual hallucination detection.

\textit{Language model} approaches exhibit divergent patterns. FActScore achieves competitive AP scores in certain languages (e.g., Finnish, French) but consistently shows poor IoU performance, suggesting it identifies hallucinated regions but struggles with precise span boundaries. \texttt{Qwen2.5-32B-it} with 3-shot prompting demonstrates language-specific strengths, achieving the best AP for German and Chinese, though its IoU scores remain moderate.

\begin{takeawaybox}
Fine-tuned multilingual encoder models consistently outperform both uncertainty-based and LLM-based approaches. But the gap between AP and IoU metrics across all methods indicates that precise span-level boundary detection remains a significant challenge.
\end{takeawaybox}

\subsection{Cross-lingual Transfer}
\label{sec:transfer}
To evaluate the cross-lingual transferability of PsiloQA, we compare mmBERT-base models trained on the complete multilingual PsiloQA dataset against models trained on individual language subsets. We evaluate performance on both PsiloQA and Mu-SHROOM test sets to measure within-distribution and cross-dataset generalization.

Table~\ref{tab:cross_ling} demonstrates that training on the full multilingual PsiloQA dataset enables robust cross-lingual transfer. The multilingual model consistently outperforms language-specific models across most target languages, with improvements observed even for languages with distinct scripts (Arabic, Hindi) and those from different language families.

\begin{takeawaybox}
Multilingual training on PsiloQA enables superior cross-lingual transfer compared to language-specific training, with benefits extending across different scripts and language families.
\end{takeawaybox}

\subsection{Knowledge Transfer}\label{sec:knowledge_transfer}
To evaluate generalization across datasets, we assess model transferability~\cite{karpov2023knowledge} by comparing models fine-tuned on synthetic PsiloQA versus human-annotated RAGTruth. We utilize \texttt{mmBERT-base} as our strongest baseline. We fine-tune it in three configurations: (i) exclusively on RAGTruth\textsubscript{QA}, (ii) solely on PsiloQA\textsubscript{en}, and (iii) on both datasets combined. All configurations use identical hyperparameters and are evaluated on three independent benchmarks: FAVA-Bench, HalluEntity, and Mu-SHROOM\textsubscript{en}.

Table~\ref{tab:ragtruth_vs_psilo} reveals that PsiloQA\textsubscript{en}-trained models consistently outperform RAGTruth\textsubscript{QA} across benchmarks. Most notably, PsiloQA achieves substantial gains on Mu-SHROOM\textsubscript{en} -- representing a 45\% improvement in IoU. Similar advantages appear on HalluEntity, while FAVA-Bench shows limited performance across all configurations, suggesting valuable task differences. Combining both datasets demonstrates selective benefits, with PsiloQA-based configurations (alone or combined) dominating. Joint training achieves the highest AP on HalluEntity. On Mu-SHROOM\textsubscript{en}, the combined model maintains strong performance, ranking second only to PsiloQA alone.

The advantage of PsiloQA\textsubscript{en} might stem from its larger training set size, yet PsiloQA's synthetic generation remains more than 17 times cheaper than manually curated RAGTruth\textsubscript{QA} (Section~\ref{sec:psiloqa}).

\begin{takeawaybox}
The superior ability of PsiloQA to transfer knowledge could be attributed to its larger size compared to RAGTruth, yet its generation cost is significantly lower than the labeling cost of RAGTruth. 
\end{takeawaybox}

\section*{Conclusion}
In this work, we introduced PsiloQA, a large-scale, multilingual, span-level hallucination detection dataset constructed using a scalable and cost-effective pipeline. Our approach leverages real hallucinations produced by LLMs in a zero-context setting and employs GPT-4o for automated span-level annotations. PsiloQA offers extensive language coverage and supports diverse LLM architectures, providing a valuable resource for evaluating and training hallucination detection models.

Through comprehensive evaluations across multiple baselines -- including uncertainty quantification, encoder-based detectors, and LLM-based methods -- we demonstrated the effectiveness of PsiloQA for benchmarking hallucination detection. Our results reveal that fine-tuned encoder-based methods, particularly multilingual models like mmBERT and ModernBERT, outperform other baselines, though significant challenges remain. Additionally, we showed that PsiloQA supports strong cross-lingual generalization and outperforms human-annotated datasets like RAGTruth in knowledge transfer experiments, despite being over 17 times cheaper to produce.

Our findings highlight the feasibility and advantages of using synthetically generated datasets with automated high-quality annotations to improve the robustness and factuality of LLMs. Future work will explore extending the PsiloQA pipeline to other generation tasks, such as summarization and data-to-text generation, further broadening its utility in hallucination research.

\section*{Limitations}
While PsiloQA presents significant advancements in span-level hallucination detection across languages, several limitations remain: 

\noindent \textbf{Annotation Source Bias}: PsiloQA relies exclusively on GPT-4o for both generating question–answer pairs and annotating hallucination spans. This introduces potential bias in annotation and generation patterns, as the judgment of a single model may not reflect broader consensus or generalize well across diverse use cases. This bias could be substantially mitigated by using an ensemble of annotators composed of several state-of-the-art models with span averaging. We consider this a promising direction for future work. 

\noindent \textbf{Task Narrowness}: The current version of PsiloQA is limited to the question-answering (QA) task. While QA is a strong proxy for factual reasoning, other generative tasks such as summarization, dialogue, and data-to-text generation also suffer from hallucinations and warrant similar treatment.

\noindent \textbf{Hallucination Type Coverage}: Unlike datasets that inject controlled hallucination types (e.g., FAVA), PsiloQA does not explicitly cover a diverse taxonomy of hallucinations. The hallucinations in PsiloQA arise naturally from LLM errors in a zero-context setting, which may result in skewed distributions and underrepresentation of certain error types.

\noindent \textbf{Language Resource Imbalance}: Despite covering 14 languages, the sample distribution across languages is uneven, and lower-resource languages may suffer from fewer high-quality examples. Additionally, many baselines used for comparison are predominantly trained or optimized for English, potentially underestimating performance in other languages.

\noindent \textbf{Dependency on Wikipedia}: Using Wikipedia as the sole source of context limits the topical, stylistic, and cultural diversity of the dataset. While Wikipedia provides clean, factual content across many languages, its coverage is uneven: some languages, cultures, and topics are better represented than others, potentially introducing cultural or regional biases into the dataset. Consequently, models trained on this data may inherit these biases. Moreover, real-world applications often involve noisier or domain-specific data.

\subsection*{Ethical Considerations}
This work involves the creation and analysis of a multilingual question-answering (QA) dataset, PsiloQA, designed for evaluating span-level hallucination detection in large language models (LLMs). We address several ethical aspects to ensure responsible data generation, annotation, and usage:

\noindent \textbf{Data Source and Privacy}: All questions and corresponding answers were generated using publicly available information from Wikipedia. The dataset contains no personally identifiable information (PII), and no sensitive or private data was collected, stored, or processed at any stage.

\noindent \textbf{Intended Use and Limitations}: PsiloQA and the associated models are intended solely for research purposes, particularly in the development of more trustworthy and interpretable QA systems. While our models aid in detecting hallucinations, they are not error-free and should not be considered reliable for deployment in high-stakes or real-time decision-making systems (e.g., healthcare, legal domains) without rigorous domain-specific evaluation and validation.

\noindent \textbf{Fairness and Inclusivity}: The dataset includes samples across 14 languages to support multilingual research. However, language coverage is uneven, and performance disparities may exist due to varying model training resources and language-specific complexities. Researchers should account for these disparities when interpreting results or deploying tools based on PsiloQA.

\noindent \textbf{Avoidance of Misuse}: We explicitly discourage the use of our dataset or trained models for surveillance, censorship, or automated moderation without human oversight. The tools are not intended for identifying or suppressing content and must not be used to enforce ideologically biased or discriminatory practices.

\noindent \textbf{Transparency and Reproducibility}: We provide complete documentation of our data generation and annotation pipeline, including prompt designs and filtering mechanisms, to ensure transparency and reproducibility. Our approach emphasizes the use of real hallucinations generated by LLMs in a zero-context setting, promoting authentic error analysis.

\noindent \textbf{Model Dependency and Bias}: Since both the generation and annotation of PsiloQA rely on GPT-4o, there is an inherent risk of model bias influencing the dataset. Although GPT-4o was among the state-of-the-art models available during dataset development, its judgments may reflect underlying model biases or fail to align with human consensus in edge cases. Furthermore, GPT-4o’s proficiency varies across languages, which may affect the consistency and quality of cross-lingual annotations. Future iterations of PsiloQA may incorporate diverse model perspectives and human-in-the-loop validation to mitigate this concern.

By making PsiloQA publicly available, we aim to support the development of robust, multilingual hallucination detection systems while promoting ethical, fair, and responsible AI research.

\bibliography{custom}

\appendix

\section{License and Infrastructure}
Experiments utilized 2 NVIDIA A100 GPUs, totaling around 50 GPU-hours. Models were used according to their licenses: Qwen 2.5 under Apache 2.0. We release our dataset under CC-BY-4.0.

\section{Packages}
To generate the PsiloQA dataset, we used the VLLM~\cite{vllm} package for efficient inference of LLMs on available GPUs, following the default hyperparameters recommended in the Hugging Face README files. Encoder model training was performed using the Transformers library~\cite{transformers}, with training hyperparameters detailed in Appendix~\ref{app:encoderhyper}.

\section{Encoder Hyperparameters}\label{app:encoderhyper}
\begin{table}[!htb]
\centering
\small
\begin{tabular}{@{}lc@{}}
\toprule
\textbf{Hyperparameter}       & \textbf{Value} \\ \midrule
learning\_rate                & 1e-5           \\
num\_train\_epochs            & 6              \\
weight\_decay                 & 0.01            \\
% gradient\_accumulation\_steps & 40             \\
batch\_size                   & 8              \\ 
\bottomrule
\end{tabular}
\caption{Training followed LettuceDetect~\cite{lettucedetect} hyperparameters for six epochs, with the best validation checkpoint selected.}\label{tab:hyperparams-encoder}
\label{tab:encoderhyper}
\end{table}

\onecolumn

\newpage
\section{PsiloQA Dataset Statistics}\label{app:psiloqa-dataset-statistics}
\begin{table}[hbt!]
\small
\resizebox{\textwidth}{!}{%
\begin{tabular}{@{}clcccc@{}}
\toprule
\textbf{Language} & \textbf{Language Model}                               & \textbf{\# of parameters} & \textbf{Avg \# of spans} & \textbf{Avg span length} & \textbf{\# of samples} \\ \midrule
ar                & SeaLLMs/SeaLLM-7B-v2.5                     & 7-9B                      & 1.35                     & 9.09                     & 2072                   \\
ca                & occiglot/occiglot-7b-es-en-instruct        & 7-9B                      & 1.07                     & 11.18                    & 6240                   \\
cs                & mistralai/Mistral-7B-Instruct-v0.3         & 7-9B                      & 1.60                     & 16.18                    & 4984                   \\
de                & malteos/bloom-6b4-clp-german-oasst-v0.1    & 3-7B                      & 0.97                     & 9.96                     & 1378                   \\
en                & HuggingFaceH4/zephyr-7b-beta               & 7-9B                      & 2.41                     & 22.46                    & 665                    \\
en                & HuggingFaceTB/SmolLM2-1.7B-Instruct        & 1-3B                      & 2.20                     & 21.36                    & 608                    \\
en                & HuggingFaceTB/SmolLM2-135M-Instruct        & <1B                       & 2.09                     & 36.80                    & 581                    \\
en                & HuggingFaceTB/SmolLM2-360M-Instruct        & <1B                       & 1.85                     & 25.97                    & 578                    \\
en                & ServiceNow-AI/Apriel-5B-Instruct           & 3-7B                      & 1.37                     & 10.66                    & 3525                   \\
en                & TinyLlama/TinyLlama-1.1B-Chat-v1.0         & 1-3B                      & 1.87                     & 21.46                    & 2151                   \\
en                & tiiuae/falcon-7b-instruct                  & 7-9B                      & 1.70                     & 14.96                    & 1595                   \\
en                & togethercomputer/Pythia-Chat-Base-7B-v0.16 & 7-9B                      & 1.29                     & 9.31                     & 2042                   \\
es                & Iker/Llama-3-Instruct-Neurona-8b-v2        & 7-9B                      & 1.43                     & 14.84                    & 2364                   \\
eu                & google/gemma-7b-it                         & 7-9B                      & 1.03                     & 10.81                    & 3853                   \\
fa                & Qwen/Qwen2-7B-Instruct                     & 7-9B                      & 1.22                     & 7.46                     & 4550                   \\
fi                & BSC-LT/salamandra-7b                       & 7-9B                      & 0.83                     & 2.37                     & 4512                   \\
fi                & Finnish-NLP/llama-7b-finnish-instruct-v0.2 & 7-9B                      & 1.09                     & 9.38                     & 2561                   \\
fr                & croissantllm/CroissantLLMChat-v0.1         & 1-3B                      & 1.56                     & 26.84                    & 2026                   \\
hi                & google/gemma-7b-it                         & 7-9B                      & 1.03                     & 10.81                    & 3853                   \\
hi                & nickmalhotra/ProjectIndus                  & 1-3B                      & 1.25                     & 25.65                    & 1801                   \\
hi                & sarvamai/sarvam-1                          & 1-3B                      & 0.98                     & 11.67                    & 3331                   \\
it                & sapienzanlp/modello-italia-9b              & 7-9B                      & 1.39                     & 15.13                    & 2181                   \\
sv                & utter-project/EuroLLM-9B-Instruct          & 7-9B                      & 1.12                     & 11.27                    & 7729                   \\
zh                & Qwen/Qwen2-7B-Instruct                     & 7-9B                      & 1.22                     & 7.46                     & 4550                   \\
zh                & Qwen/Qwen2.5-3B-Instruct                   & 3-7B                      & 1.11                     & 1.46                     & 1170                   \\
zh                & ikala/bloom-zh-3b-chat                     & 3-7B                      & 1.28                     & 3.04                     & 3309                   \\ \bottomrule
\end{tabular}
}
\label{tab:general-stats}
\caption{The list includes the utilized LLMs along with their corresponding languages and statistics like the average number of spans and the average span length.}
\end{table}

\input{psiloqa_stats}

\clearpage

\newpage
\section{PsiloQA Dataset Sample}\label{app:psiloqa-dataset-sample}
\begin{table*}[!htb]
\small
\resizebox{\textwidth}{!}{%
\begin{tabular}{p{3.5cm} p{2.5cm} c p{2.5cm} p{3.5cm} }
\toprule
\textbf{Passage}                                                                                                                                                                                           & \textbf{Question}                                                    & \textbf{Complexity} & \textbf{Golden Answer}          & \textbf{LLM Answer}                                                                 \\ \midrule
Lillian Richter (1915–2000) was an American lithographer. Richter did work for the Works Progress Administration (WPA).                                                                                    & When was Lillian Richter born?                                       & Easy                & 1915                            & Lillian Richter was born in \hl{1932}.                                               \\ \midrule

Things I Carry Around is the eleventh studio album by Australian country music artist Troy Cassar-Daley.                                                                                                   & What is the title of the eleventh studio album by Troy Cassar-Daley? & Easy                & Things I Carry Around           & The eleventh studio album by Troy Cassar-Daley is "\hl{The Greatest Hits}"           \\ \midrule

Albert Herman (1887–1958) was an American actor, screenwriter and film director. Herman was a prolific director, working mainly on low-budget movies for companies such as Producers Releasing Corporation & For which company did Albert Herman mainly work as a director?       & Hard                & Producers Releasing Corporation & Albert Herman mainly worked as a director for the company \hl{Albert Herman \& Cie}. \\ \midrule

Edmond Armand René Thorailler, né le 17 juin 1911 à Nogent-le-Roi (Eure-et-Loir) et mort le 24 août 2004 à Notre-Dame-de-Monts (Vendée), est un notaire et homme politique français. & Quelle est la date de naissance d'Edmond Thorailler? & Easy & 17 juin 1911 & \hl{1923}.  \\ \midrule

Bloomfield Road est le nom d'un stade de football localisé à Blackpool, en Angleterre.
C'est l'enceinte du club principal de la ville, le Blackpool Football Club, depuis 1901. Ce stade de 16 220 places a été inauguré le 28 octobre 1899. & Quel est le nom du stade situé à Blackpool, en Angleterre? & Easy & Bloomfield Road & \hl{Blackpool Stadium}  \\ \midrule

Der Rotstielige Zwerg-Täubling  oder Rotstieliger Birken-Täubling (Russula font-queri) ist ein Pilz aus der Familie der Täublingsverwandten. Es ist ein seltener Täubling und typischer Birkenbegleiter mit kupferrotem Hut und leicht rosa überlaufenem Stiel. Sein Sporenpulver ist gelblich. & Wie heißt der Rotstielige Zwerg-Täubling auf Latein? & Easy & Russula font-queri & \hl{Amanita rubescens-Team}  \\

\bottomrule
\end{tabular}
}
\caption{Example of PsiloQA samples. Each sample contains the passage retrieved from Wikipedia, the question and the Golden Answer of some complexity level generated by GPT-4o, the LLM answer generated by some LLM, and the hallucination ranges (highlighted in \hl{red}) annotated by GPT-4o by comparing the LLM answer with the Golden Answer.}
\end{table*}

\newpage
\section{QA Pairs Generation Prompt}\label{app:psiloqa-generation-prompts}
\begin{figure*}[!htb]
\begin{small}
% \begin{mdframed}
\hrulefill
\begin{tcbverbatim}
Generate 3 question-answer pairs with different levels of complexity (easy, medium and hard).
You must only create questions that require knowledge of the passage.
Format your answer as a Python list with 3 jsons.
Each json should contain "question", "answer" and "complexity" fields.

Do not generate questions that require imagination.
The questions should be factual, so do not generate questions that ask for subjective opinion or reasoning.
It should be enough to know the facts from the provided passage.
Each question must contain exactly one question.
Do not make any reference to the passage in the question-answer pair.

Use the language in which the passage is written.

{answer_length_constraint}

**Passage:** {p}
\end{tcbverbatim}
\hrulefill
\end{small}
\caption{Prompt for question-answer pairs generation used for PsiloQA creation. We ask GPT-4o to generate three different question-answer pairs of different complexity using the retrieved passage from Wikipedia. We also control the length of the answer. In 33\% of the cases we ask GPT-4o to generate long and detailed answers.}\label{fig:qa-generation-prompt}
\end{figure*}

\newpage
\section{Inconsistency Detection Prompt}\label{app:inconsistency-detection-prompt}
\begin{figure*}[!htb]
\begin{small}
% \begin{mdframed}
\hrulefill
\begin{tcbverbatim}
Your task is to find any inconsistencies with the correct information in LLM's answer.
Carefully read the user's question, the golden answer, the relevant passage, and LLM's answer.
The relevant passage contains information for answering the question.
LLM did not see the relevant passage while generating the response.

General instructions:
- If LLM refused to answer the question, then answer by simply copying LLM's answer. There is no inconsistency if LLM did not provided any answer to the question.
- If LLM's answer does not contain information relevant to the question, and the information does not contradict the relevant passage, then answer by simply copying LLM's answer.
- If LLM's answer is relevant to the question, use the opening tag [HAL] and closing tag [/HAL] to highlight areas of inconsistency with the golden answer and relevant passage information. Inconsistency means something that is related to the topic of the question, but contradicts the relevant passage or introduces new information.
- If LLM's answer is consistent with the golden answer, do not highlight anything, answer by simply copying LLM's answer.

How to highlight spans:
- Each span could contain from 1 to several words. In rare, specific cases, it could be longer.
- Make spans as precise as possible, do not highlight the entire answer.
- Highlight spans at the word level, not the character level.

DO NOT ADD ANY CHANGES TO LLM'S ANSWER EXCEPT [HAL] and [/HAL]!

Begin your answer with "**Highlighted LLM Response:**".
\end{tcbverbatim}
\hrulefill
\end{small}
\caption{Prompt for inconsistencies detection in LLM answers. We pass to prompt questions, answers of different LLMs and the golden answers generated by GPT-4o. In the prompt, we ask GPT-4o to find any spans of inconsistencies in the golden answer. Here we consider the obtained inconsistencies with the gold and the context as hallucinations.}\label{fig:inconsistency-detection-prompt}
\end{figure*}

\newpage
\section{Filtering Prompts}\label{app:filtering-prompts}
\begin{figure*}[!htb]
\begin{small}
% \begin{mdframed}
\hrulefill
\begin{tcbverbatim}
Classify the following question into one of three categories:
1. INCOMPLETE_QUESTION
Incomplete questions that use pronouns with no clear antecedent ("Where is THIS located?" / "What is HE famous for?").
Questions that refer to a list/paragraph/excerpt that isn't included in the question itself.

2. SUBJECTIVE
Questions whose answers require value judgments or a subjective opinion.

3. NORMAL
Any question with a clear subject ("Where is Jacksonville located?" / "Who is Alexander Montenegro?").
The subject may be ambiguous or polysemous, but it must not be impersonal.
Open-world questions that name their subject (even if broad or multi-answer) are NORMAL.

Instructions:
Output only one of the three categories: INCOMPLETE_QUESTION, SUBJECTIVE, or NORMAL.

Examples:
INCOMPLETE_QUESTION
Q: What were the set scores in the final match?
A: INCOMPLETE_QUESTION
Q: Which religious order was the abbey associated with?
A: INCOMPLETE_QUESTION
Q: What is the chemical formula of perrhenic acid as stated in the passage?
A: INCOMPLETE_QUESTION

SUBJECTIVE
Q: What are some of the primary aims of the Athletics Club Lechia Gdansk?
A: SUBJECTIVE
Q: What is the significance of the Fahey-Murray ministry in the context of New South Wales government history?
A: SUBJECTIVE

NORMAL
Q: What was the population of Hazleton according to the 2020 census?
A: NORMAL
Q: What role did William Arrington hold in the Illinois State Senate between 1955 and 1973?
A: NORMAL
Q: Which Olympic Games did Lee Jun-ho represent South Korea in?
A: NORMAL
Q: How many league appearances did James Ryan make in the EFL?
A: NORMAL
\end{tcbverbatim}
\hrulefill
\end{small}
\caption{Prompt for the detection of subjective and incomplete questions. Subjective questions require a personal opinion rather than a factual answer. Incomplete questions lack a clear subject and are therefore unanswerable.}\label{fig:questions-filtering}
\end{figure*}
\begin{figure*}[!htb]
\begin{small}
\hrulefill
\begin{tcbverbatim}
You are a strict detector of 'I don't know' style answers.
Given an answer string, return only the word TRUE if the answer expresses refusal to answer due to lack of information, or otherwise clearly states that it cannot answer. 
If the model provides any meaningful information regarding the topic or makes assumptions, return only the word FALSE, even if there was uncertainty anywhere in the answer. 
Do not explain.
\end{tcbverbatim}
\hrulefill
\end{small}
\caption{Prompt for detecting cases when LLM refuses to answer.}\label{fig:answers-filtering}
\end{figure*}

\newpage
\section{LLM Baseline Evaluation Prompt}\label{app:baseline-prompt}
\begin{figure*}[!htb]
\begin{small}
% \begin{mdframed}
\hrulefill
\begin{tcbverbatim}
Please act as an objective error detector. You will be given the user's question, a relevant passage, and the LLM's response.  
Your task is to analyze the response to the question and identify the parts of the response that are most likely to contain errors or inconsistencies.  
Taking into account the question and the answer provided by the language model, you need to wrap erroneous words/phrases in the answer with special tokens:  [HAL] erroneous word/phrase  [/HAL].  
Before the question, you will be given a passage - a factual reference that will help you identify factual errors. Use it as a guide when highlighting hallucinations - mark words/phrases as hallucinations if they do not correspond to the information in the passage.

Example {x3}:
Knowledge source: {sampled passage (optional)}
Question: {sampled question}
Answer: {sampled answer}
Answer with highlighted spans: {sampled highlighted answer}

Knowledge source: {passage}
Question: {question}
Answer: {answer}
Answer with highlighted spans:
\end{tcbverbatim}
\hrulefill
\end{small}
\caption{Prompt for evaluating the baseline model \texttt{Qwen2.5-32B-instruct} in English.  
We evaluate baseline in 3-shot mode, with examples sampled from the PsiloQA validation set.  
The prompt is translated for each language, and the few-shot examples are picked from the corresponding language.}\label{fig:baseline-prompt}
\end{figure*}

\end{document}